\documentclass[10pt,twocolumn,letterpaper]{article}

\usepackage{iccv}
\usepackage{times}
\usepackage{epsfig}
\usepackage{graphicx}
\usepackage{amsmath}
\usepackage{amssymb}
\usepackage{booktabs}
\usepackage{multirow}
\usepackage{tabularx}

\usepackage[breaklinks=true,bookmarks=false]{hyperref}

\iccvfinalcopy % *** Uncomment this line for the final submission

\def\iccvPaperID{3896} % *** Enter the ICCV Paper ID here

\ificcvfinal\fi

\begin{document}

%%%%%%%%% TITLE
\title{3D Face Arbitrary Style Transfer}

\author{Xiangwen Deng , Yingshuang Zou, Yuanhao Cai, Chendong Zhao, Yang Liu  \\ 
Zhifang Liu, Yuxiao Liu, Jiawei Zhou, Haoqian Wang \\ % \thanks{Corresponding author: wanghaoqian@tsinghua.edu.cn}\\
Shenzhen International Graduate School, Tsinghua University, China\\
% {\tt\small \textsuperscript{1}   \{dengxw22,zouys22,liu-yang22,liuzf20,liuyuxia22,zhoujw22\}@mails.tsinghua.edu.cn}\\
% {\tt\small \textsuperscript{3}   \{wanghaoqian\}@tsinghua.edu.cn}
% {\tt\small \textsuperscript{2}   \{cd896614\}@gmails.com}\\
% For a paper whose authors are all at the same institution,
% omit the following lines up until the closing ``}''.
% Additional authors and addresses can be added with ``\and'',
% just like the second author.
% To save space, use either the email address or home page, not both
}

\maketitle
% Remove page # from the first page of camera-ready.
\ificcvfinal\thispagestyle{empty}\fi

%%%%%%%%% ABSTRACT
\begin{abstract}
Style transfer of 3D faces has gained more and more attention. However, previous methods mainly use images of artistic faces for style transfer while ignoring arbitrary style images such as abstract paintings. To solve this problem, we propose a novel method, namely Face-guided Dual Style Transfer (FDST). To begin with, FDST employs a 3D decoupling module to separate facial geometry and texture. Then we propose a style fusion strategy for facial geometry. Subsequently, we design an optimization-based DDSG mechanism for textures that can guide the style transfer by two style images. Besides the normal style image input, DDSG can utilize the original face input as another style input as the face prior. By this means, high-quality face arbitrary style transfer results can be obtained. Furthermore, FDST can be applied in many downstream tasks, including region-controllable style transfer, high-fidelity face texture reconstruction, large-pose face reconstruction, and artistic face reconstruction. Comprehensive quantitative and qualitative results show that our method can achieve comparable performance. All source codes and pre-trained weights will be released to the public.
\end{abstract}

\begin{figure}[htbp]
  \centering
  \includegraphics[width=0.45\textwidth]{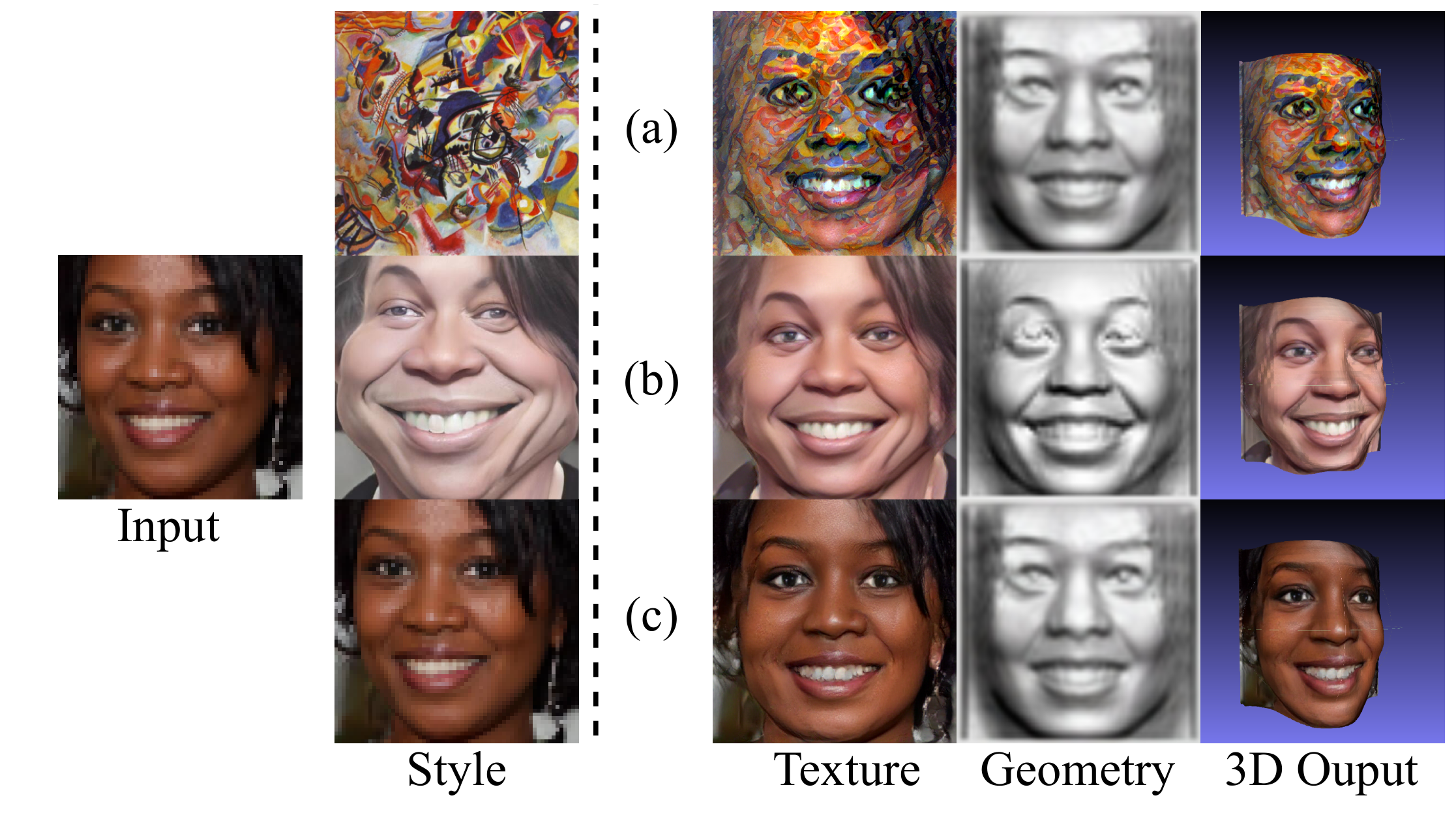}
  \caption{The goal of 3D face style transfer is to transfer the style of an example-guided image to texture or geometry. Given a face image as input, our method supports three tasks. (a) We support 3D face style transfer using arbitrary abstract images. (b) We support style transfer using the artistic face-style image. Our geometry can also be transferred according to the input face style. (c) We can achieve high-fidelity texture reconstruction by transferring the original input as style to a low-quality reconstructed texture. }
  \label{fig:shouye}
\end{figure}

% \begin{figure}[htbp]
 %  \centering
 %  \includegraphics[width=0.45\textwidth]{first_image_2.pdf}
 %  \caption{The goal of 3D face reconstruction is to transfer the style of an example-guided image to texture or geometry. Given a face image as input, our method supports three tasks. (a) We can achieve high-fidelity texture reconstruction by transferring the original input as style to a low-quality reconstructed texture. (b) We support 3D face style transfer using abstract images. Clear facial features can be maintained even under large style weights. (c) We support style transfer using the artistic face-style image. Our geometry can also be transferred according to the input face style.}
%   \label{fig:shouye}
% \end{figure}

%%%%%%%%% BODY TEXT
\section{Introduction}
\label{sec:intro}

Face style transfer has been widely studied in the 2D research field \cite{AniGAN,carigan,warpgan,jojogan,blendgan}, but there is little related work in the 3D vision domain. As shown in Fig.\ref{fig:shouye}, the goal of 3D face style transfer is to transfer the style of an example-guided image to texture or geometry. Compared with 2D, 3D face style transfer has the advantage of multi-view consistency \cite{fenerf,ide3d,FaceBlit}. With the rise of the metaverse and digital human, 3D face style transfer has wide applications in face animation modeling and style editing \cite{exemplar,Caricatureshop,3dcarigan,FaceBlit}.

\begin{figure}[htp]
  \centering
  \includegraphics[width=0.48\textwidth]{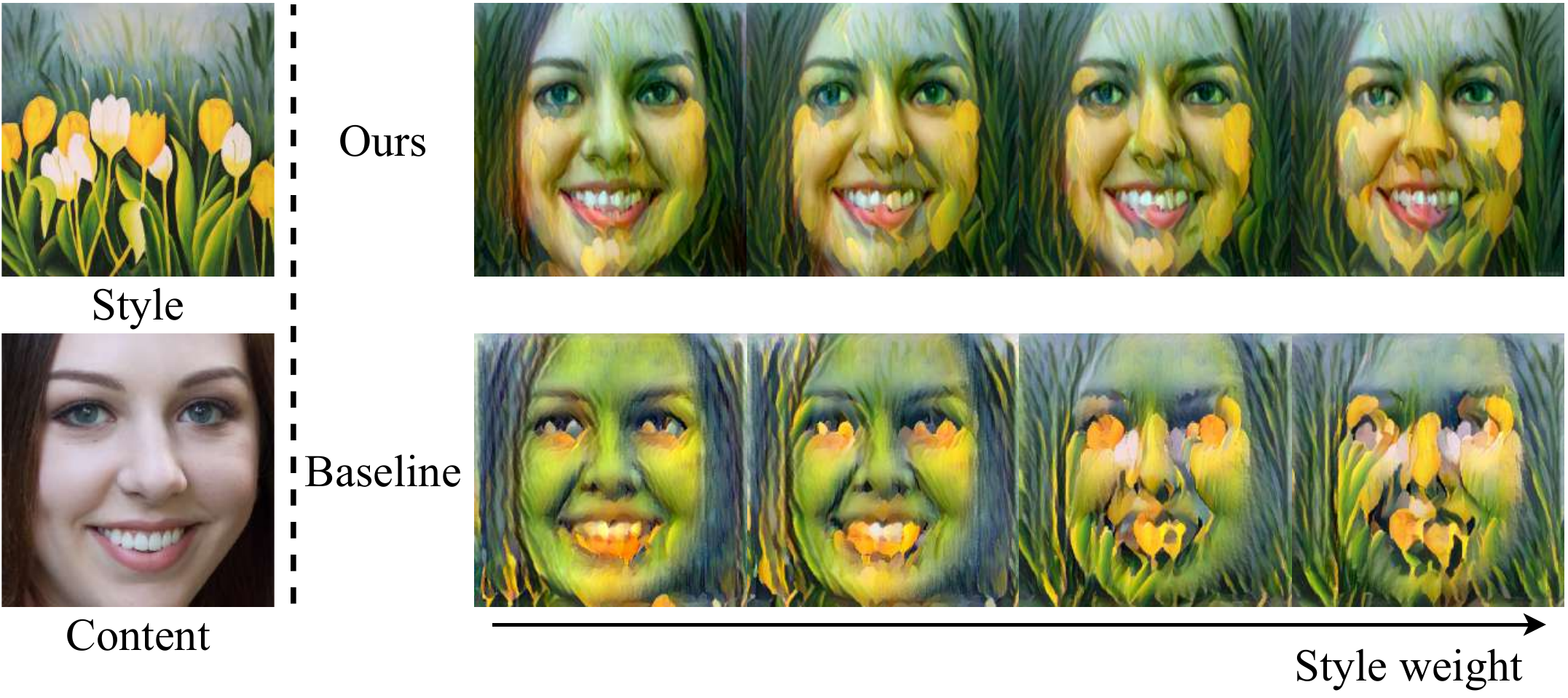}
  \vspace{-2mm}
  \caption{Compared with the previous face transfer method \cite{gatys2016,kolkin,wct2}, FDST can obtain better quality style transfer results. When using the large-weight style transfer, our method can preserve the shape of the original face while the baseline~\cite{kolkin} collapses.}
  \label{fig:duibi2}
  \vspace{-3.5mm}
\end{figure}

% 这里需要再精炼简化
3D face style transfer is different from 2D face style transfer because the style of geometry and texture need to be considered separately. For face geometry style transfer, our goal is to derive the geometry after the fusion of the two styles. current methods such as \cite{exemplar, Caricatureshop,3dcarigan} are modeled with FLAME \cite{FLAME} and the geometry is transferred via the landmark warp algorithm \cite{carigan,warpgan}. This scheme works for caricature images but fails in Van Gogh-type abstract art image style transfer because facial keypoints cannot be detected.
For face texture style transfer, the common method is StyleGAN \cite{style2}. However, due to the limitation of StyleGAN inversion space, arbitrary style transfer ($e.g.,$ abstract paintings) cannot be achieved. Other methods which can realize arbitrary style transfer are neural style transfer \cite{gatys2015,Johnson,Fujun,gatys2016}. Among them, STROTSS \cite{kolkin} has achieved impressive results and can transfer style textures well. But it also collapses when transferring arbitrary style images to faces. In summary, traditional VGG-based style transfer methods suffer from two issues as depicted in the lower line of Fig.\ref{fig:duibi2}. On the one hand, when low-weight style transfer is used, the style cannot be effectively transferred. On the other hand, when high-weight style transfer is exploited, the texture details are severely distorted and the original face shape cannot be preserved. In this case, the face shape collapses and cannot be aligned with the geometry of the face, resulting in poor rendering results.

% In addition, there has been a surge of interest in high-fidelity face texture reconstruction field \cite{Ganfifit,kim2021learning,zhang2022learning} in recent years. Until now, high-fidelity face texture reconstruction tasks are mainly divided into methods based on GAN training and StyleGAN iteration \cite{zhang2022learning,gan2shape, Liftingstylegan}. GAN-based methods require datasets, but they are not open source. For the methods using StyleGAN, they use the strategy of inversion, which can reconstruct the face in high resolution. However, due to the limitation of inversion space, this method cannot effectively learn abstract art faces.

% 启发（在introduction里面不用说这个) ：
To cope with the above problems, we propose a novel method, namely Face-guided Dual Style Transfer (FDST), for 3D face arbitrary style transfer. First, FDST employs a 3D decoupling module to separate facial geometry and texture. For the geometry, we propose a style fusion strategy. For the texture, we design an optimization-based module, dubbed Dual Disentangled Style Guidance (DDSG). Specifically, DDSG is guided by two style images. Besides the normal-style image input, we take the raw face input as another style input to enhance the face prior. In this way, DDSG module allows our style transfer to keep texture details even under large style weights, as shown in the upper line of Fig.\ref{fig:duibi2}. In addition, our FDST can also be applied in many downstream tasks, including high-fidelity face texture reconstruction, region-controllable style transfer, large-pose face reconstruction, and artistic face reconstruction. Especially in the high-fidelity face texture reconstruction task, our method can effectively reconstruct both photo-realistic face texture and abstract artistic face texture. Different from general deep learning-based face high-fidelity texture restoration algorithms that require a large-scale dataset for training, our FDST does not need to be trained and only requires a high-resolution face style image for inference.

To sum up, our contributions are listed as follows:  % 不改了, 就这些贡献

\begin{itemize}
\vspace{-1.5mm}
\item We propose a novel framework FDST for 3D face arbitrary style transfer. To the best of our knowledge, this is the first attempt for 3D face ``arbitrary" style transfer.%which supports geometry or texture style transfer separately with more geometric and texture details.
\vspace{-1.5mm}
\item We design an optimization-based face texture transfer module  DDSG. By using a dual style-guided strategy, DDSG can jointly preserve the face shape and transfer the texture details  even using large style weights.
%\textcolor{red}{cyh: I am here.}
\vspace{-1.5mm}
\item Comprehensive experiments demonstrate that our FDST achieves state-of-the-art (SOTA) results. Besides, the applications in downstream tasks like high-fidelity face texture reconstruction tasks, region-controllable style transfer, large-pose face reconstruction, and artistic face reconstruction also suggest the practical values of our algorithm. To the best of our knowledge, our FDST is also the first style transfer-based high-fidelity face texture reconstruction method.
\end{itemize}

%------------------------------------------------------------------------

\section{Related Work}
% 修改中间的逻辑, 是否正常阐述流派的发展
% 修改最后的逻辑和对于我们方法的对比. 

\subsection{ 2D Arbitrary Style Transfer }
% 想得到的优点：我们同时支持迁移两次, 精度和两次相当, 而且更加鲁棒, 不容易形变. 时间更少, 
2D style transfer aims to modify the style of the input image while preserving the content of the input image \cite{Johnson, Fujun, gatys2016}. Gatys \emph{et al.}~\cite{gatys2015} first propose a neural style transfer method based on an optimization strategy. Many later works are inspired by it, such as SANet \cite{sanet}, AdaIN \cite{adain}, WCT \cite{wct2} and BlendGAN \cite{blendgan}. However, these methods only transfer the reference style to the source image without considering the local semantic style, leading to visual artifacts in the output. To solve this problem, some methods are proposed to improve the transfer quality based on semantic self-similarity \cite{kolkin}. Yet, this self-similarity mechanism is not effective to preserve the texture details of faces when using arbitrary style guidance. As shown in Fig.\ref{fig:duibi2}, when the large style weight transfer is used, the facial texture and facial features are distorted. As a result, the face can not be aligned, leading to low-quality 3d rendering results.

\begin{figure*}[tbp]   
  \centering
  \includegraphics[width=1.0\textwidth]{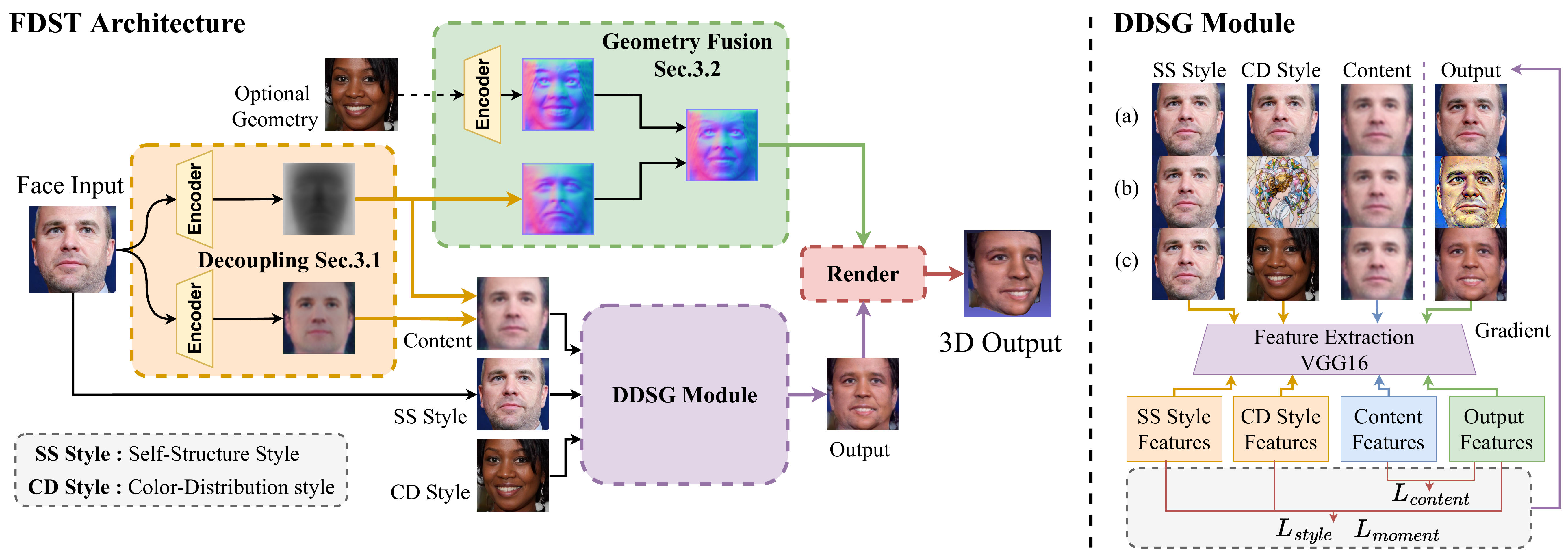}
  \caption{FDST (left) is mainly divided into three modules: the 3D decoupling module, the geometry fusion module, and the texture transfer module $\mathcal{DDSG}$. Inputting a face image, our method can decouple geometry and texture for style transfer separately. After style transfer, we render the geometry and texture to get the 3D output.  $\mathcal{DDSG}$ (Right) supports three tasks: (a)  high-fidelity texture reconstruction. (b) arbitrary style transfer which deforms texture only. (c) face style transfer which deforms both texture and geometry. These tasks differ only in the CD style image, they all use the input face image as the SS style image to enhance to detail of texture.}
  \label{frame}
\end{figure*}

\subsection{ 3D Face Style Transfer}% 这里是我们的重点, 所以需要多写一点
%我们的优点：
% exampler是最直接的对手, 但是他是基于参数化模型, 虽然可以扭曲, 但是不能很好的还原人脸的细节褶皱, 有平均脸的倾向. 
The goal of 3D face style transfer is to use a 2D style image to guide a 3D face to deform its geometry and texture. This field can currently be divided into two categories. The first category relies  on StyleGAN such as EG3D \cite{eg3d}, Ide-3d \cite{ide3d}, Cips-3d \cite{cips-3d}, Your3dEmoji \cite{Your3dEmoji0}. These methods use StyleGAN interpolation \cite{toonify, jojogan} for style transfer. But since the inversion method of StyleGAN can only encode faces, it cannot achieve style transfer of arbitrary images. The second category is 3D stylization methods based on parametric template modeling and neural rendering. These methods usually generalize neural style transfer to 3D grids \cite{H-Kato,Paparazzi}. Yet, these methods do not consider the semantic information of the face. A recent work \cite{exemplar} is modeling by FLAME \cite{FLAME} and using STROTSS \cite{kolkin} for texture style transfer rendering. This scheme works for caricature images but fails in Van Gogh type abstract art style transfer because the facial keypoints cannot be detected for morphing. Thus, the second category of methods can not achieve high-quality style transfer for arbitrary style images.

\subsection{High-Fidelity 3D Face Texture Reconstruction } % 确实我们做的任务也不算是补全任务, 而是增强任务和zhang的那个stylegan的方法非常像. 
% 我们的优点：可以arbitrary图像增强, 包括抽象风格人脸, 艺术人脸, 迭代速度更快, 
Many current 3D face reconstruction methods cannot reconstruct face texture details \cite{Ostec,animatable}. So high-fidelity 3D face texture reconstruction has always been a hot topic \cite{Ganfifit,Learning-detailed,Synthesizing}. The methods can be divided into two categories: modeling-based methods and GAN-based image generation methods. Firstly, the typical modeling-based method is 3DMM \cite{guo2020towards,tran2019towards}. 3DMM is constructed from a small number of face scans. Due to the linear and low-dimensional nature of the model, it is difficult to capture high-frequency details, resulting in blurry face textures. 
Secondly, GAN-based image generation methods \cite{Nonlinear-3d,Gun-Hee,uvgan} are also competitive methods to improve texture quality. Iterative methods \cite{zhang2022learning,gan2shape, Ostec} based on StyleGAN optimization can produce the most realistic face UV images. However, they require a large-scale dataset to train and are time-consuming. For example, GANFIT \cite{Ganfifit} takes 30 seconds to generate a UV image of the input face while OSTeC \cite{Ostec} takes 5 minutes.  In contrast, we only need 70 seconds to infer a single image without training. More importantly, previous methods are unsuitable for the arbitrary face ($, e.g., $ abstract face image) reconstruction due to the limitation of the training dataset.

%%%%%%%%%%%%%%%%%%%%%%%%%%%%%%%%%%%%%%%%%%%%%%%%%%%%%%%%%%%%%%%%%%%%%%%%%%%%%%%%%%

\section{Method}

In this section, we first describe the Face-guided Dual Style Transfer method (FDST).  Our style transfer framework supports three tasks: (a) high-fidelity texture reconstruction, (b) arbitrary style transfer which deforms texture only, and (c) face style transfer  deforming both texture and geometry. These tasks differ only in style images. They all use the input face image as one of the style images to enhance texture details.

FDST is depicted in Fig.\ref{frame}. Given a face image as the input, we first need to obtain the face texture image and depth image through a pre-trained 3D-decoupling module (Sec.\ref{sec:Decoupling}). Then we use a geometric style fusion module (Sec.\ref{sec:Geometry}) to reconstruct geometry. If the style image has the geometry of the face, we can use the optional geometry module to get the fusion geometry. Next, we elaborate on our key contribution, the optimization-based texture style transfer module $\mathcal{DDSG}$ (Sec.\ref{sec:Texture}). Finally, the geometry and texture are rendered \cite{H-Kato} to produce the 3D output.

\subsection{3D Face Decoupling} \label{sec:Decoupling}
Our goal is to decompose the input face image into its corresponding geometric depth image and texture canonical image, which can be used as the input for subsequent style transfer tasks. To this end, we use the pre-trained model of Unsup3D \cite{unsup3d} since it has been widely applied in unsupervised face reconstruction \cite{zhang2022learning,gan2shape,Liftingstylegan,zhang2022physically} in recent years. It uses encoders to disentangle a facial image $I$ into intrinsic factors$ (d, a, \omega, l)$, including a depth map $d \in R_+$, an albedo image $a \in R^3$, a directional $l \in S^2$, and a viewpoint $\omega \in  R^6 $, where $d$, $a$, and $l$ are in canonical space. Then, with these factors, the canonical image $C$ is reconstructed by lighting function $\Lambda$, $a$, $d$ and $l$ as follows:
\begin{equation}
  C = \Lambda (a, d, l),
  \label{equation:C1}
\end{equation}
where the canonical image $C$ will be used as materials for texture transfer later and the depth image $d$ will serve as the material for geometry transfer. We directly use the pre-trained Unsup3D encoder to predict the depth image $d$ and canonical image $C$. It must be pointed out that since Unsup3D depends on the high quality of training data, it cannot handle extreme facial expressions, large poses, \emph{etc.} Meanwhile, the texture of the image reconstructed by Unsup3D is still blurry for the HD face input image. We will describe the $\mathcal{DDSG}$ module in Sec.\ref{sec:Texture} to address these issues.

%%%%%%%%%%%%%%%%%%%%%%%-------------ok

% \begin{figure}[tbp]   
%   \centering
%   \includegraphics[width=0.5\textwidth]{style_transfer(3)_compressed.pdf}
%   \caption{This is our texture transfer module $\mathcal{DDSG}$ (Sec. \ref{sec:Texture}). Our method is based on iterative optimization. Firstly, we use the Laplacian eigenvalues of the content image to initialize the output image. Then, we extract features from the output image, the content, the SS style image, and the CD style image through VGG. Finally, we calculate the corresponding loss and gradient and then optimize the output image iteratively. }
%   \label{styletransfer}
%  \end{figure}

\subsection{Geometry Style Transfer} \label{sec:Geometry}
According to the types of target style images, there are two categories of face style transfer. The first category transfers abstract styles ($e.g.,$ abstract landscapes) that do not contain the geometric information of faces. Another category transfers face-like images that contain geometric information of faces. For the first category, we directly reconstruct the geometry mesh from the previously decoupled depth image $d$. For the second category, we can also use encoder \cite{unsup3d} to get the depth image of the style face. Then we use the fusion strategy of depth image to achieve the geometric style transfer. By this means, we rebuild the geometry mesh. Since the depth images of the content face and the style face are center-aligned, we can directly use the interpolation fusion strategy to achieve the purpose of geometric style transfer, as follows:
\begin{equation}
d_{out}=\alpha \cdot d+(1-\alpha) \cdot d', % 这里的融合应该也是取决于style1和style2的比例
\label{equation:d}
\end{equation}
where $\alpha\in (0,1)$ controls the stylization effect, $d_{out}$ denotes the final fusion output depth image, $d$ indicates the output depth of the first image, and $d'$ represents the output depth of the second image.

%%%%%%%%%%%%%%%%%%%%%%%-------------ok
\subsection{Texture Style Transfer} \label{sec:Texture}
% 重构逻辑了：

Our goal is to transfer textures with high quality without causing distortion. Since we need to use the texture image reconstructed by Unsup3D as the content image for style transfer, the low-quality Unsup3D texture will affect the final effect (Sec. \ref{sec:Decoupling}). Therefore, our texture transfer needs to learn the texture features of style images and enhance the texture details of human faces.

Our key insight is that the face priors can be used as one of the styles to help style transfer avoid blurred and distorted facial features and shapes especially when using large transfer weights  for face arbitrary style transfer.

Considering the structure of texture and the color distribution of arbitrary style, two style images are used as guidance. Specifically, according to different roles, these two styles are named as the Self-Structure style (SS style) and the Color-Distribution style (CD style). SS style guidance is used to obtain textural and facial structure information from the original input face. Since our $\mathcal{DDSG}$ module is based on self-similarity, the SS style guidance of the raw input image can be used as face prior to improve texture detail quality. CD style guidance obtains texture color information according to different types of tasks. As Shown in Fig. \ref{frame}, this is the first time to obtain different cues from two style images in one style transfer process. In this way, we decouple the sources of style color and structure. Thus, we derive more clues to improve the texture details of the output.

%%%%%%%%%%%%%%%%%%%%%%%-------------ok
%%%%%%%%%%%%%%%%%%%%%%%-------------ok

\noindent\textbf{Style loss.} The style loss $L_{style}$ is computed between a content image $C$ and two style images ($S_1$ and $S_2$). We use the same feature extractor as \cite{exemplar}. Firstly, we feed two style images and one content image into a pre-trained VGG16 network and extract their multi-layer feature maps. Bilinear upsampling is exploited to match the spatial resolution. Then the feature maps are concatenated along the channel dimension. This produces a hypercolumn at each pixel, which includes low-level features capturing edges and colors and high-level features capturing semantics. We denote $A = \{ A_1, ..., A_k\} $ as the set of $k$ feature vectors of the content image $C$, $B = \{B_1, ..., B_k\}$ as a set of $k$ feature vectors of style image $S_1$, $B' = \{B'_1, ..., B'_k\}$ as the set of $k$ feature vectors of style image $S_2$, where $k$ is the number of pixels. The style loss is formulated from the Earth Movers Distance (EMD) \cite{emd} as follows:
\begin{small}
\begin{equation}
  \begin{aligned}
  L_{style}(A, &S_1, S_2) =  \alpha EMD(A, B)+ \beta EMD(A, B')\\ &=   \min_{T\geq0} \sum_{ij} T^1_{ij} Cost_{ij} + \min_{T\geq0} \sum_{ij} T^2_{ij} Cost_{ij}, \\
  &s.t. \sum_{j}T_{ij}=i/k \quad and \quad  \sum_{j}Cost_{ij}=i/k,
  \label{equation:EMD}
  \end{aligned}
  \end{equation}
\end{small}
\noindent where $T^1$ and $T^2$ are the "transfer matrix" that defines the partial pairwise assignment between pixel $i$ in $A$ and pixel $j$ in $B$ or $B'$. We use $\alpha$ and $\beta$  to control the degree of style mixing in S1 and S2, respectively. In practice, we set $\alpha = 1$, $\beta = 3$. The $Cost_{ij}$ is the "cost matrix" which measures the feature cosine distance $\cos (A_i, B_j)$ between each pixel in $A$ and each pixel in $B$ as follows:
\begin{equation} 
  C_{ij}=D_{\cos}(A_i,B_j)=1-\frac{A_i\cdot B_j}{\left\lVert A_i \right\rVert \left\lVert B_i \right\rVert}.
%Lm=1/d||μA-μB_2||%+1/d^2||ΣA-ΣB_1||
\label{equation:C_{ij}}
\end{equation}
It is time-consuming to directly optimize the EMD. So we adopt the suggestion of \cite{exemplar,kolkin} to relax it.

\noindent\textbf{Moment loss.} The cosine distance we used above ignores the size of the feature vector. In practice, this will lead to visual artifacts in the output (see our Supplement). To solve this problem, we construct a moment-matching loss:
\begin{equation} 
  L_{moment}=\frac{1}{d}(||\mu_A-\mu_{B'}||+||\Sigma_A-\Sigma_{B}||), 
%Lm=1/d||μA-μB_2||%+1/d^2||ΣA-ΣB_1||
\label{equation:Lm}
\end{equation}
% 其中,µA,ΣA为集合A中特征向量的均值和协方差,而µB1和ΣB1则是B1中的特征向量的均值和协方差,而µB2和ΣB2则是B2中的特征向量的均值和协方差. 这样设置的原因是因为两个输入的风格功能不同. style1负责为纹理几何细节, 所以使用ΣB1衡量,而style2负责为捕获纹理色彩辅助, 所以用μB2衡量. 
where $\mu_A$, $\Sigma_{A}$ are mean and covariance of feature vectors in set $A$. $\mu_{B'}$ is the mean of feature vectors in $B'$. $\Sigma_{B}$ is the covariance of feature vectors in $B$. The reason for this setting is that the goals of the two style inputs are different. The SS style aims to capture the structure of texture, so it is measured by $\Sigma_{B}$. While The CD style is responsible for capturing texture color, so it is measured by $\mu_{B'}$.

\begin{figure*}[tbp]
  \centering
  \includegraphics[width=0.95\textwidth]{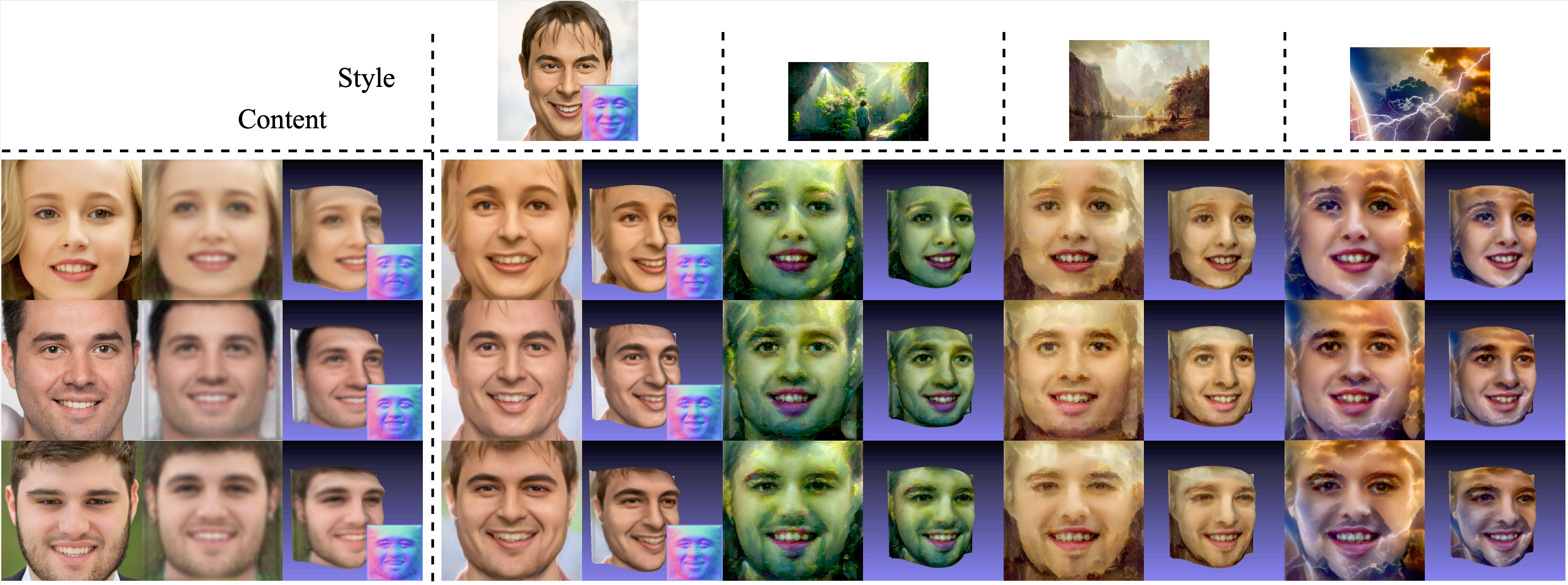}
  \caption{Results of our arbitrary style transfer. Our method can transfer the arbitrary style while maintaining the identity of the face. }
  \label{fig:fig4}
\end{figure*}
% If the style image is an abstract art image, we just style transfer the texture. If the style image is a face-like art image, we will transfer both geometry and texture.

\noindent\textbf{Content loss.} Similar to the style loss, we use VGG to obtain features for both content image $C$ and the generated image $X$. We denote $M = \{M_1, ..., M_k\}$ as the set of $k$ feature vectors of $X$. $N = \{N_1, ..., N_k\}$ is denoted as the set of $k$ feature vectors of $C$. And $k$ is the number of pixels. The content loss is defined as the difference between the self-similarity of the two feature maps:
\begin{small}
\begin{equation} 
% Lcontent(X, C)  = 1/n^2 Σ (||cos(Ai,Aj) − cos(Bi,Bj)||)
L_{content}(X, C) =\frac{1}{n^2} \sum_{ij}||\cos(M_i,M_j)-\cos(N_i,N_j)||,
\label{equation:Lcontent}
\end{equation}
\end{small}
where $X$ is the generated image and $C$ is the content image. In other words, the cosine distance computed between the feature vectors of the content image and the generated image should remain unchanged. Because self-similarity preserves semantic and spatial layout while allowing pixel values in $X$ to be significantly different from those in $C$.
%%%%%%%%%%%%%%%%%%%%%%%-------------ok

\noindent\textbf{Total loss.} The overall training objective is formulated as a weighted combination of the style loss, moment loss, and content loss:
\begin{equation} 
L_{texture} = \lambda  L_{style} + \eta  L_{content} + L_{moment}
\label{equation:Ltexture}
\end{equation}
where $\lambda  $ controls the style weight and $\eta $ controls the content weight. In practice, we set $\lambda = 0.5$, $\eta = 1$

%%%%%%%%%%%%%%%%%%%%%%%%%%%%%%%%%%%%%%%%%%%%%%%%%%%%%%%%%%%%%%%%%%%%%%%%%%%%%
\begin{table}[tbp]
  \centering
  \resizebox{1\columnwidth}{!}{
  \begin{tabular}{cccccc}
      \toprule
      Method       & Ours    & AdaIN   & NST     & STROTSS & WCT2     \\
      \midrule
      Average rank & 1.275 & 3.400 & 3.072 & 3.341 & 3.911  \\
      \bottomrule
  \end{tabular}
  }
  \caption{Perceptual study result analysis.}
  \label{users}
\end{table}

\section{ Experiments} 
% \subsection{Datasets}
% FEI \cite{} and FFHQ \cite{} datasets are used for texture quantitative analysis. FEI contains images of the number of faces and angles, which we use to measure our probation effect. FFHQ includes image of front faces, which we use to measure our effect on high resolution front faces. See our supplementary materials for details.

\subsection{Implement Details}
% 逻辑完备
For the 3D decoupling module, we use the pre-trained Unsup3D \cite{unsup3d} encoders to predict the depth image $d$ and canonical image $C$. For the $\mathcal{DDSG}$ module, we use the same neural style transfer iteration strategy as \cite{gatys2016,kolkin}. Specifically, the output image needs to be iteratively processed under four spatial resolutions, including 64$\times$64, 128$\times$128, 256$\times$256, and 512$\times$512. At each resolution, the output image, content image, SS style image, and CD style image are in the form of a Gaussian pyramid. Then they are fed into VGG network to derive feature maps. These features are used to calculate the total loss. 200 iterations are required at each resolution to minimize the objective loss function. We use the RMSprop strategy to update the entry of the output image in the pyramid.

%计算loss的方法是对于输出图像的每一层我们用vgg提取特征后, 与content, style1和style2构成的金字塔提取的每层特征（他们也同样处理) 来一起最小化我们的目标函数 Ltexture . 

%%%%%%%%%%%%%%%%%%%%%%%%%%%%%%%%%%%%%%%%%%%%%%%%%%%%%%%%%%%%%%%%%%%%%%%%%%%%%%

% 目标是证明: 说白了就是我们和他们效果一样,而且还具有很多其他的优势. 包括用户评价和更多应用的展示请见我们的supplement. 

% 这里就是常见的一般意义上风格迁移和人脸重建,我们对于原始的unsup3d和strotss的优越性. 加上和对于其他方法的可比性或者优越性. 

\begin{figure*}[tbp]
  \centering
  \includegraphics[width=0.9\textwidth]{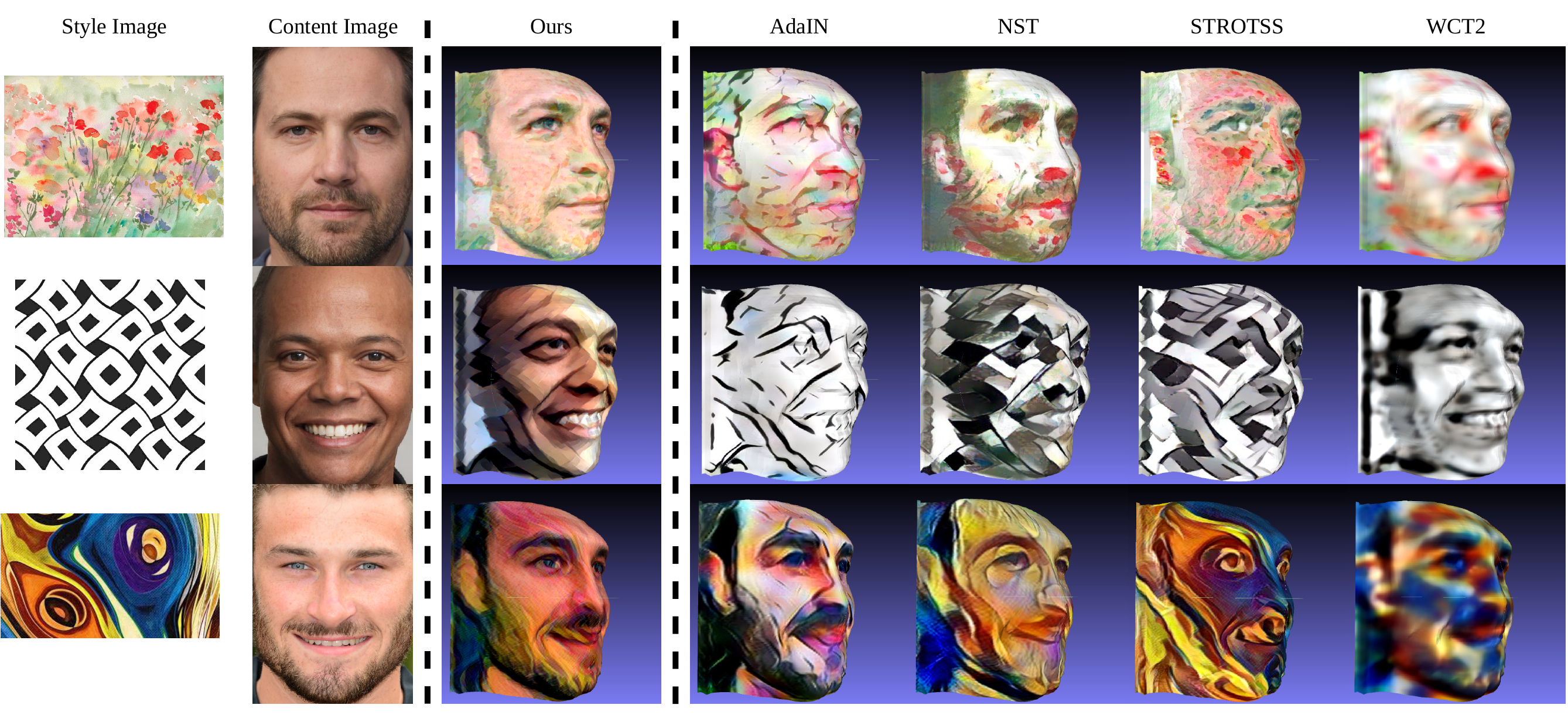}
  \caption{Comparison results with baseline arbitrary style transfer methods. We compare with neural style transfer-based methods including AdaIN \cite{adain}, NST \cite{gatys2016}, STROTSS \cite{kolkin}, WCT2 \cite{wct2}. Because our method preserves the clearest facial texture (eyes, mouths, noses), our method obtains the best result of 3D output.}
  \label{fig:contrast}
\end{figure*}

\begin{figure}[tbp]
  % 我们的方法支持区域mask引导的风格迁移的结果。我们指定区域进行风格迁移，而其他区域不变，如最右边钢铁侠的鼻子等。
  \centering
 \includegraphics[width=0.45\textwidth]{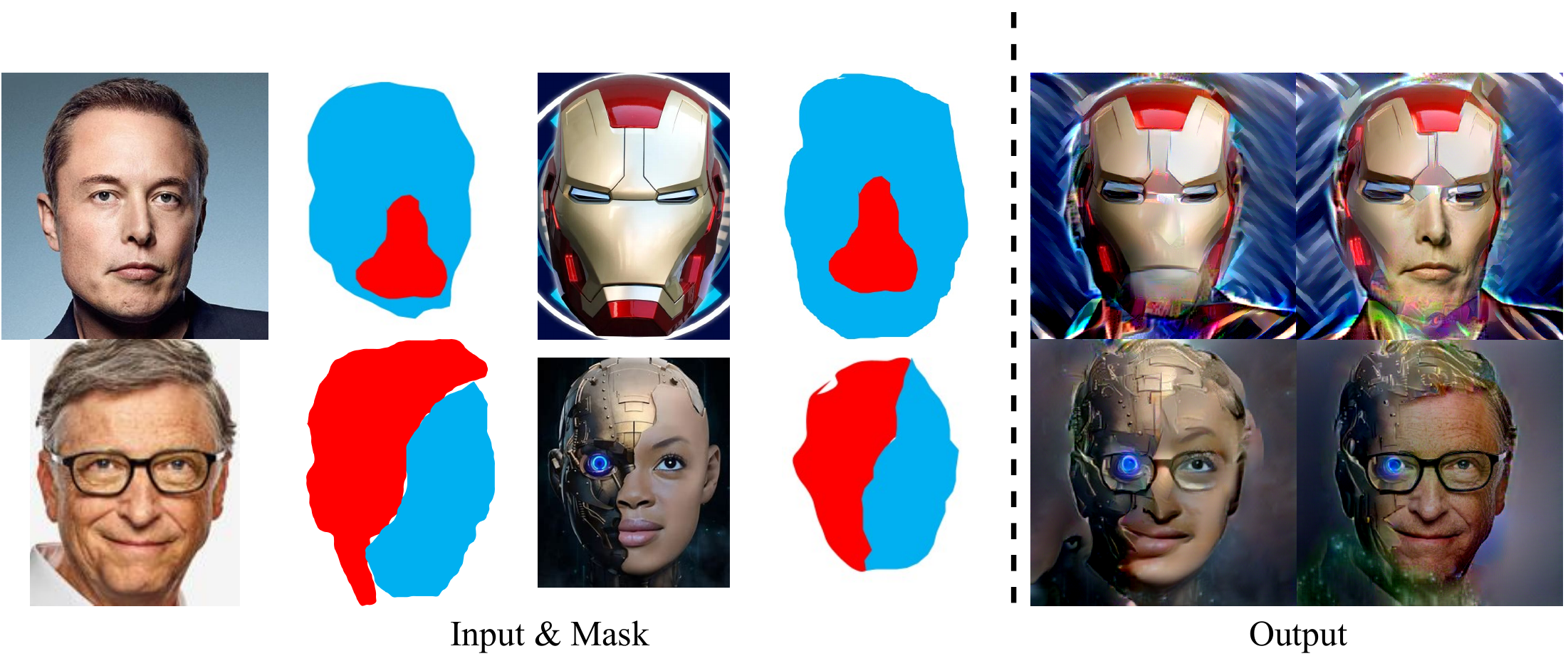}
   \caption{Our method supports the region-mask guided style transfer. We can select specific regions for style transfer and keep other regions unchanged.}
   \label{fig:point}
  \end{figure}

\subsection{Main Results}
As shown in Fig. \ref{fig:fig4}, we show the visual results of our arbitrary style transfer. For images with geometric style, our FDST can transfer geometry style, otherwise only transfer texture. In the following part, we describe the comparison between our FDST and baseline and quantitative evaluation.

\noindent\textbf{Comparsion with Baseline.} We compare our method with the baseline method in 3D face style transfer. Because our method uses arbitrary style images, we compare our method with classical methods of arbitrary style transfer. These methods include AdaIN \cite{adain}, NST \cite{gatys2016}, STROTSS \cite{kolkin}, WCT2 \cite{wct2}.
 
As shown in Fig. \ref{fig:contrast}, our FDST achieves high-quality 3D face transfer results.  Since some style images have no geometric information, we can only transfer the texture style without deforming the geometry. Therefore, to ensure the quality of the rendering results, we need to preserve the shape of the face ($e.g.,$ eyes, mouth, nose) during the style transfer process. Our method treats the input face image as the SS style, optimized along with the CD style image. Because $\mathcal{DDSG}$ is based on the self-similarity mechanism, the benefit of this design is that more input texture information can be obtained. If the style image is an abstract face that has the geometric style of the face, our FDST can both deform the geometry and texture. (see Sec. \ref{sec:Geometry}).

\noindent\textbf{Quantitative Analysis.} For the task of 3D face style transfer, it is necessary to effectively preserve the structure of the facial features. Otherwise, excessive style transfer will severely distort the facial shape, which will lead to poor rendering results. However, evaluating style transfer methods is a challenging task since style transfer has no definitive "Ground Truth". Conducting a perceptual user study is the most common way to evaluate different style transfer methods. As listed in Tab. \ref{users}, we take a similar approach as \cite{nstreview,reshuffle}. Specifically, we use ten groups of images. Each group contains a comparison of our method and four other methods, including AdaIN \cite{adain}, NST \cite{gatys2016}, STROTSS \cite{kolkin}, WCT2 \cite{wct2}. The outputs of each method are a face albedo image and a 3D render frontal face image. All images from each group were presented side by side in random order. The 30 different test users, aged 20 to 40, were randomly recruited online. Users had unlimited time to rate their preference from 1 to 5 (1 being the best, and 5 being the worst). We then use the average score of each method as an evaluation metric to measure the quality of style transfer. Overall, our method is the most popular compared to other methods. please refer to the supplementary for more details.

\subsection{Application}

\subsubsection{ Regional Controllable Style Transfer.} Our $\mathcal{DDSG}$ can support regional guided style transfer. The results is shown in Fig. \ref{fig:point}. Similar to STROTSS \cite{kolkin}, by using mask guidance for the input image and the style image separately, $\mathcal{DDSG}$ can choose the region of the corresponding color for style transfer. Furthermore, $\mathcal{DDSG}$ also supports style transfer with different weights for different regions. Thus, it is possible to transfer only the required region to the specified target region, while the style of other regions remains unchanged.

\subsubsection{High Fidelity Texture Reconstruction}
\begin{figure*}[tbp]
  \centering
  \includegraphics[width=0.95\textwidth]{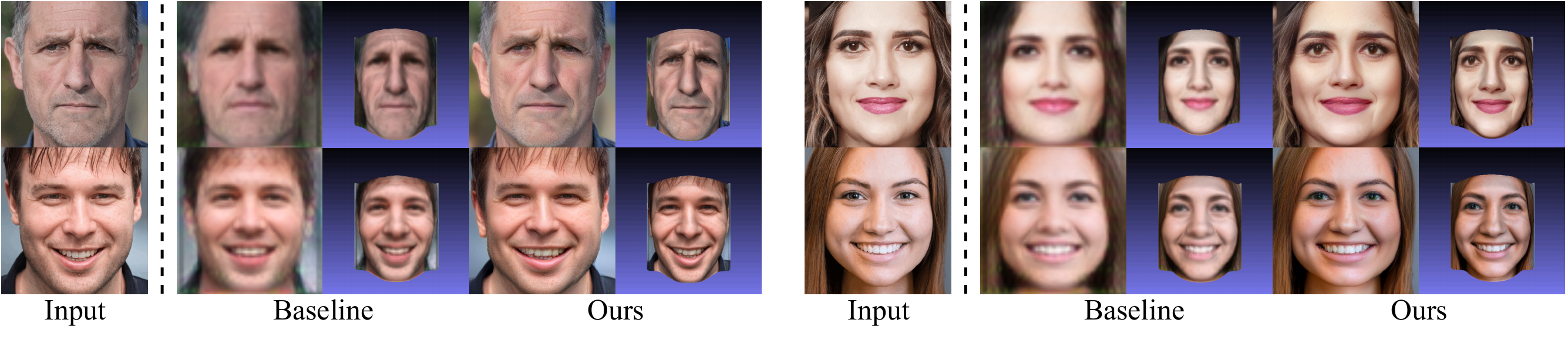}
     \caption{Results of high-fidelity face texture reconstruction. $\mathcal{DDSG}$ can use the SS style of HD input faces to enhance the blurred texture images obtained by our 3D decoupling module. Thus, the texture details of the output can be improved.}
     \label{fig:fig3}
 \end{figure*}

Thanks to the design of $\mathcal{DDSG}$, as shown in Fig. \ref{fig:fig3}, FDST can also be applied in high-fidelity face reconstruction.

\noindent\textbf{Comparsion with Baseline.} We compare our method with the recent 3D texture reconstruction methods. These methods include GANFIT \cite{Ganfifit}, Genova et al. \cite{20}, Tran et al. \cite{54}, MoFA \cite{52}.
 \begin{figure}[tbp]
  \centering
  \includegraphics[width=0.45\textwidth]{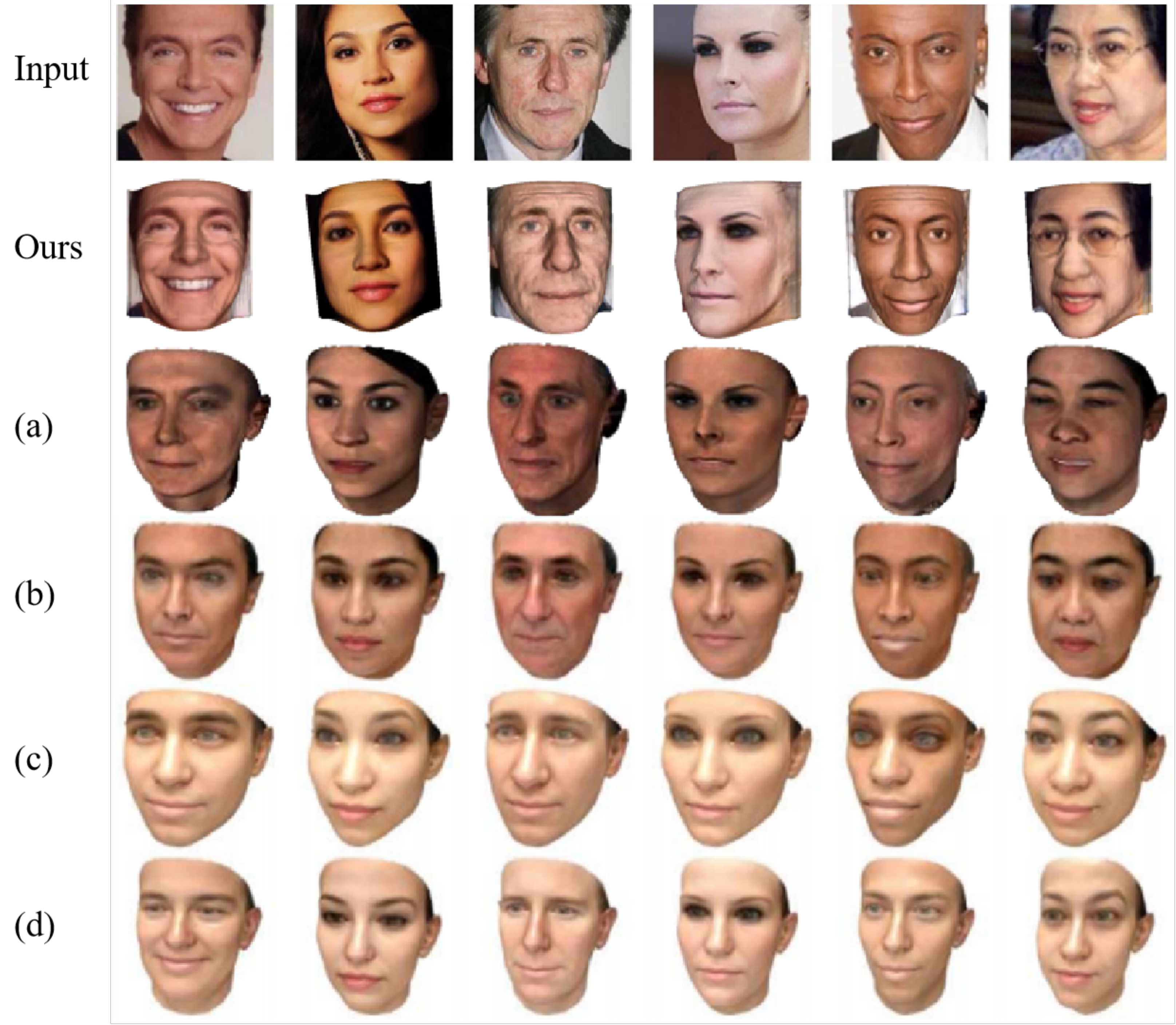}
  \caption{Comparison results with other high-fidelity face texture reconstruction methods. (a) GANFIT \cite{Ganfifit}, (b) Genova et al. \cite{20}, (c) Tran et al. \cite{54}, (d) MoFA \cite{52}. Compared to other methods, our results have better texture details, especially with the best color fidelity.}
  \label{fig:bijiao}
\end{figure}
As illustrated in Fig. \ref{fig:bijiao}, our FDST achieves comparable results on high-fidelity 3D face texture reconstruction, with even better color fidelity and texture details. Since $\mathcal{DDSG}$ relies on the self-similarity mechanism, we transfer the original high-fidelity face as styles can improve the quality of texture. In addition, our style comes from two sources: SS style and CD style. So we can control different weights to learn the texture structure and texture color in different scales. We found that assigning more weights to SS styles with frontal face inputs will also have better textures. For the input face with no canoncial front pose, we will have better performance by assigning more weights to the CD style.

 % \begin{figure*}[tbp]
  % 我们对于moment-loss的消融实验的结果。输入的input图片同时也作为SS style图片，另一个风格图片实际上作为CD style。我们展示了(a)(b)(c)(d)(e)(f)是什么的结果。可以看到只有我们的方法可以实现迁移SS风格的人脸同时保留CD style的风格。
  %  \centering
  %  \includegraphics[width=0.9\textwidth]{appendix_ablation.pdf}
  %  \caption{Results of $\mathcal{DDSG}$ ablation experiments for moment-patching  loss. Specifically, the input image is not only used as a content image, but also as an SS style image. And style images are actually used as CD style images. We show (a) using $\mu_{B'}$ and $\Sigma_{B}$. (b) using $\mu_{B}$ and $\Sigma_{B'}$. (c) moment-pathcing loss is not used. (d) Using $\mu_{B}$ and $\mu_{B'}$. (e) Use $\Sigma_{B}$ and $\Sigma_{B'}$. (f) Use $\mu_{B}$, $\mu_{B'}$, $\Sigma_{B}$ and $\Sigma_{B'}$. It can be seen that only our method (a) can transfer the face of SS style and the style of CD style at the same time.}
  %  \label{fig:appendix_ablation}
 % \end{figure*}

\noindent\textbf{Quantitative Analysis.} To evaluate the texture reconstruction quality of HD face input, we compare our methods with the baseline methods, including vanilla Unsup3D \cite{unsup3d} and Unsup3D with super-resolution DFDNet \cite{dfdnet} augmentation. We calculate PSNR, SSIM, MSE, and Cosine Similarity between the reconstructed textures and the original images. We perform our experiments on the FFHQ-mini dataset, in which we selected some of these HD frontal faces from FFHQ \cite{style2} (see our Supplement). The results of Tab. \ref{ffhq} demonstrate that our FDST has a better reconstruction quality than others.

For face frontalization evaluation, we use the FEI \cite{THOMAZ2010902} dataset which contains different rotation angles of the human face. We perform face frontalization on the images with yaw rotations of [90°, 70°, 50°, and 30°].The images with yaw rotation of 0° are regarded as the Ground Truth. And we calculate histogram similarity and average hash distance between the frontalization result and the Ground Truth image. As reported in Tab. \ref{fei}, our method significantly outperforms pixel2style2pixel \cite{pSp} and RotateAndRender \cite{RR}.

\begin{table}[tbp]
  \centering
  \resizebox{1.00\columnwidth}{!}{
  \begin{tabular}{ccccc}
      \toprule
      Method       & PSNR$\uparrow$    & SSIM$\uparrow$ & MSE$\downarrow$   & Cosine Similarity$\uparrow$     \\
      \midrule
      Unsup3D \cite{unsup3d} & 16.216 & \textbf{0.586} & 0.006 & 0.948  \\
    
      Unsup3D+DFDNet\cite{dfdnet} & 15.619 & 0.540 & 0.007 & 0.941  \\
      
      Ours & \textbf{19.143} & 0.583 & \textbf{0.003} & \textbf{0.975}   \\
      \bottomrule
  \end{tabular}
  }
  \caption{ Quality comparison of face texture reconstruction with HD input on FFHQ dataset-mini.}
  \label{ffhq}
\end{table}

\begin{table}[tbp]
\centering
\resizebox{1.00\columnwidth}{!}{
\begin{tabular}{c|cccc|cccc}
\toprule
\multirow{2}{*}{Method} & \multicolumn{4}{c}{Ahash Distance$\downarrow$}\vline  & \multicolumn{4}{c}{Hist Similarity$\uparrow$}  \\
                      & 90°     & 70°    & 50°     & 30°    & 90°     & 70°    & 50°     & 30°     \\
\midrule
R\&R \cite{RR}                 & 31.946 & 32.186 & 32.166 & 32.320 & 0.386 & 0.371 & 0.369 & 0.367  \\

pSp\cite{pSp}                     & 18.173 & 19.706 & 20.973 & 20.393 & 0.494 & 0.468 & 0.450 & 0.450  \\

Ours                    & \textbf{17.740} & \textbf{17.786} & \textbf{17.593} & \textbf{17.126} & \textbf{0.508} & \textbf{0.550} & \textbf{0.549} & \textbf{0.537} \\
\bottomrule
\end{tabular}
}
\caption{Quality comparison of face texture reconstruction with large pose input on FEI dataset.}
\label{fei}
\end{table}

\textbf{Discussion.} The task of texture reconstruction using style transfer is challenging. Because we only improve the details of the texture through style transfer, without transfer the geometry. Thus, We explain how we can ensure that the transferred texture are well-aligned to the untransferred geometry. The texture alignment relies on the self-similarity-based EMD loss (Eq. \ref{equation:EMD}). In other words, the more similar the style and content are, the better aligned transferred result we will obtain. Thus, with the help of this mechanism and SS style image, textures will be properly aligned with the geometry unless the Unsup3D geometry reconstruction is wholly failed.

Extensive experiments (Tab. \ref{ffhq}, Tab. \ref{fei}) demonstrate that our FDST can warp the input image to the output face to enhance the texture quality. In particular, unless GAN-based methods are used, previous texture reconstruction methods can only obtain blurry results even for HD input faces (Sec.\ref{fig:fig3}). Therefore, our FDST has the potential to become a new paradigm in the field of high-fidelity face texture reconstruction. Although better texture enhancement can indeed be obtained from geometric modeling, $\mathcal{DDSG}$ can directly improve the visual effects by textures enhancement, especially in areas where it is challenging to model geometry details (see Fig.\ref{fig:fig3} eye wrinkles and mustache).

%\subsubsection{Extreme 3d Face Reconstruction} 
%In addition to HD-front face reconstruction, our method also supports extreme faces, including abstract art faces and large pose faces.
 
\subsubsection{Abstract Art Face Reconstruction.} 
\begin{figure}[tbp]
  \centering
  \includegraphics[width=0.45\textwidth]{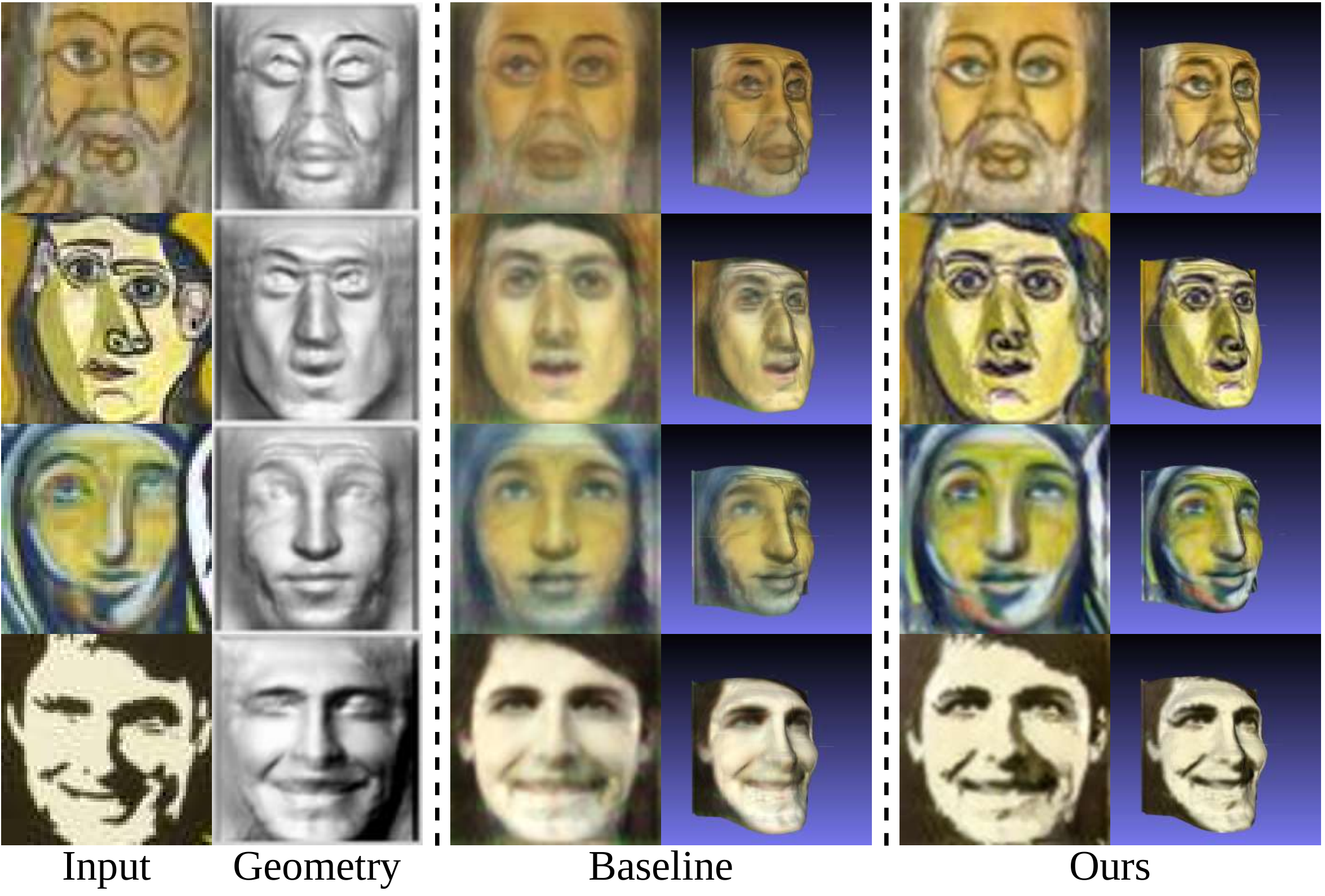}
  \caption{Results on abstract face reconstruction.}
  \label{fig:chouxiang_rencon}
\end{figure}

Our FDST can reconstruct arbitrary abstract artistic face textures as seen from Fig. \ref{fig:chouxiang_rencon}. The current abstract art texture reconstruction method is based on parametric face warp \cite{warpgan,carigan,exemplar}, which can only reconstruct the texture of abstract caricature art images and is not suitable for arbitrary abstract face images. However, because our texture reconstruction method is based on arbitrary style transfer, there is no requirement for the input face type. With only a single style image as input, the texture of arbitrary abstract faces can also be reconstructed.

\begin{figure}[tbp]
  \centering
  \includegraphics[width=0.45\textwidth]{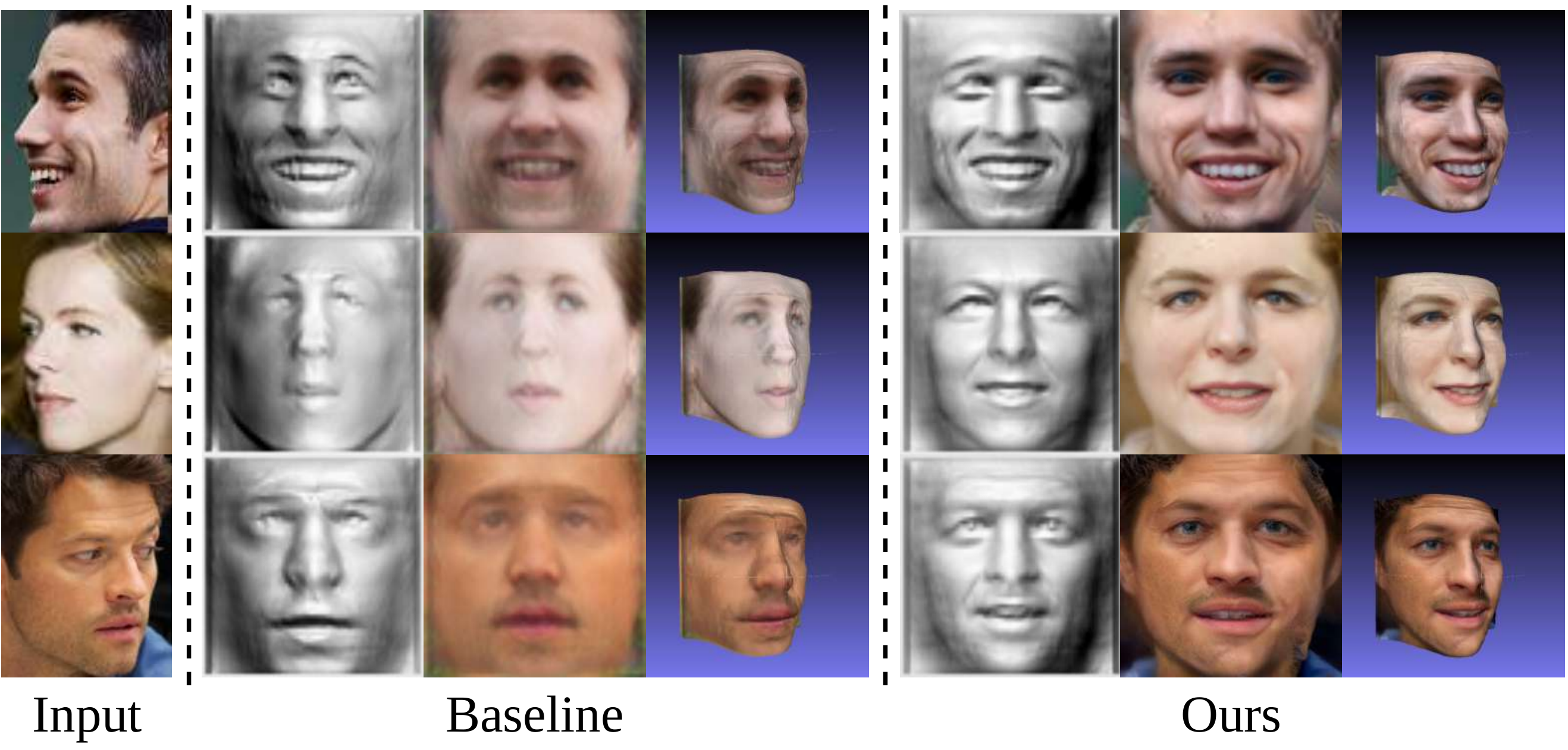}
  \caption{Results on large pose face texture reconstruction.}
  \label{fig:dazitai}
\end{figure}
 
\subsubsection{ Large-Pose Face Reconstruction.} As shown in Fig. \ref{fig:dazitai}, our method can achieve high-quality texture results for large pose images. So far, the state-of-the-art methods for large-pose face texture reconstruction are mainly based on StyleGAN. pSp \cite{pSp} is a representative work. It has high-quality texture structure results, but the color fidelity is poor. To solve this problem, we take the original input as CD style to transfer the correct color style and take the frontalizaion result of pSp as the SS style to transfer the texture structure style. (See our supplement) the frontalizaion result of pSp has the correct texture structure, the color is distorted. While the original large pose input has the correct texture color, the texture structure is distorted. Our FDST can separately get the correct texture information and color information from these two style inputs. By this means, FDST achieves high-fidelity texture transfer results for large-pose face images.

  % \subsubsection{High Fidelity Face Texture Reconstruction}
  \begin{figure}[tbp]
    \centering   
    \includegraphics[width=0.45\textwidth]{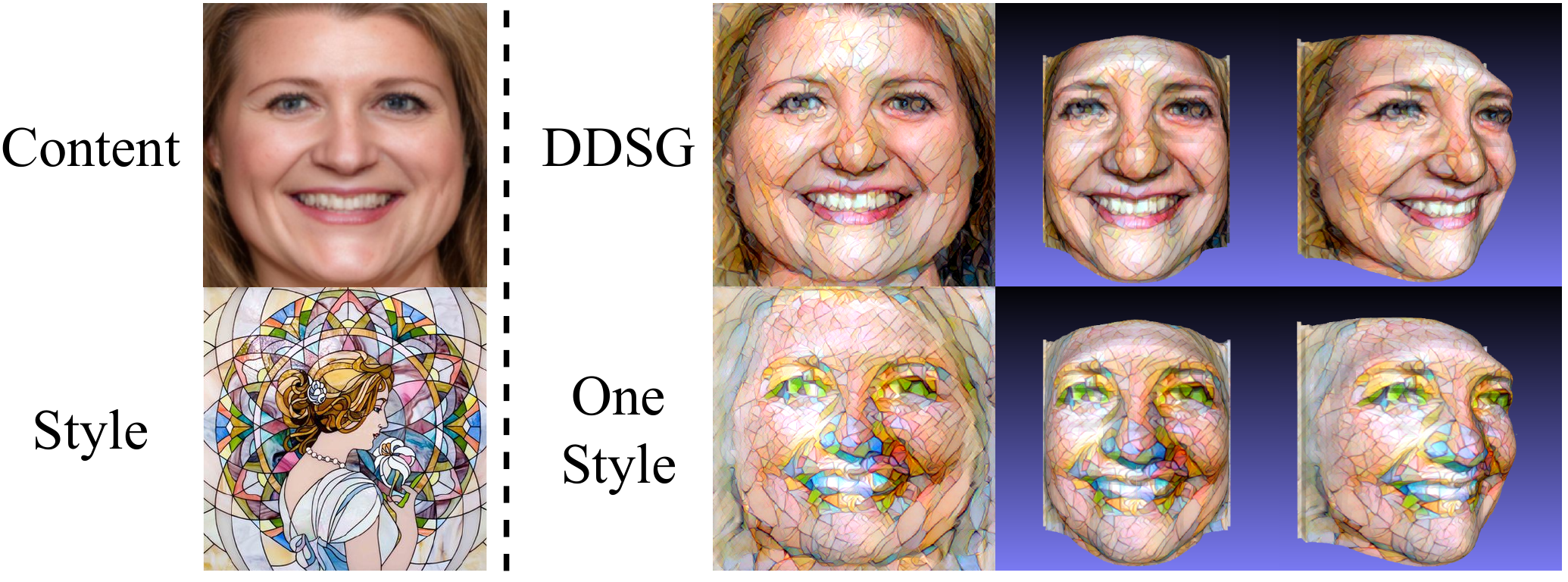}
    \caption{Ablation study results. We can see that the style transfer under single style guidance will make the facial texture deformed excessively, and the identity of the input face cannot be retained. Therefore geometry cannot be well-aligned with texture, which is not allowed in the 3D face arbitrary style transfer task. Since the geometry has stayed the same, excessive deformation in texture can lead to poor render results.}
    \label{fig:xiaorong1}
  \end{figure}
  
 % 总结,我们的方法在重建和迁移上面都取得了有竞争力的结果,更多的包括用户评价和更多应用的展示请见我们的supplement. 

%%%%%%%%%%%%%%%%%%%%%%%%%%%%%%%%%%%%%%%%%%%%%%%%%%%%%%%%%%%%%%%%%%%%%%%%%%%%%%

\subsection{Ablation Study}

\noindent{\bf One Style or Two Style.} We conduct an ablation study to compare single-style guidance and dual-style guidance (DDSG). The visual results in Fig. \ref{fig:xiaorong1} show that DDSG has better performance than single-style guidance for arbitrary style transfer. The advantage of DDSG mechanism is that it decouples the sources of styles. Specifically, the structure information can be extracted from the SS image and the color information can be captured from the CD image. Since our DDSG module is based on self-similarity, the SS style guidance of the raw input image can be used as face prior to enhance texture detail information. Compared with single style guidance, our method captures better texture color of the face under small degree style weight transfer. And our method can also better maintain the texture structure of the face under large degree style weight transfer. Otherwise, excessive style transfer will severely distort the facial texture shape, which will lead to poor 3d rendering results. Please refer to our supplement for more Ablation studies. 

% {\bf loss of $\mathcal{DDSG}$.} We evaluate our moment-patching loss through ablation experiments. To verify that the $\Sigma$ and $\mu$ designed by the moment-patching loss can decouple the style source of $\mathcal{DDSG}$, We perform ablation experiments. As shown in Fig. \ref{fig:appendix_ablation}, the $\mu_{B'}$ and $\Sigma_{B}$ of moment-patching loss adopted by our $\mathcal{DDSG}$ achieves the best style transfer results. As shown in Fig. \ref{fig:appendix_ablation} (a), the texture structure feature of the SS image can be expressed by $\Sigma$, and the color distribution feature of the CD style image can be expressed by $\mu$. On the contrary, it will leads to visual artifacts in the output if moment-patching loss is not used (Fig. \ref{fig:appendix_ablation} (c)). Furthermore, Other designs of loss failed, which will lead to distortion of the distribution of style color(Fig. \ref{fig:appendix_ablation} (b, d, e, f)).

%%%%%%%%%%%%%%%%%%%%%%%%%%%%%%%%%%%%%%%%%%%%%%%%%%%%%%%%%%%%%%%%%%%%%%%%%%%%%%
\section{Conclusions}

% \textbf{Limitations.} We obtain high-fidelity face texture reconstruction results through the style transfer method of $\mathcal{DDSG}$. Therefore, our results are dependent on the resolution of the input face image. If the input image is blurry and low resolution, Our texture reconstruction results will also be poor. We may consider using StyleGAN to enhance the transfer results of blurred images in the future.

In this paper, we propose a novel framework FDST for 3D face arbitrary style transfer. A Dual Disentangled Style Guidance (DDSG) mechanism is presented and plugged into FDST for facial texture style transfer. By using two style images as guidance, our DDSG is capable of preserving more textural details from the original input face image. Extensive experiments demonstrate that our FDST achieves state-of-the-art results. The applications in several downstream 3D face low-level vision tasks including high-fidelity face texture reconstruction tasks, region-controllable style transfer, large-pose face reconstruction, and artistic face reconstruction also reveal the practical values of our method.

%If we use input face image as style guidance, our framework is also suitable for high-fidelity face texture reconstruction. It is the first time that a style transfer method is proposed to solve the problem of face texture reconstruction. Qualitative and quantitative experiments demonstrate the effectiveness of our method. 

% {\bf limits and future works.}

{\small
\bibliographystyle{ieee_fullname}
\bibliography{egbib}
}

\end{document}